\pdfoutput=1

\documentclass[11pt]{article}

\usepackage[]{acl}

\usepackage{times}
\usepackage{latexsym}

\usepackage{graphicx}
\usepackage{picinpar}
\usepackage{subcaption}
\usepackage{colortbl}
\usepackage{booktabs}
\usepackage{multirow}
\usepackage{soul}
\usepackage{enumerate}
\usepackage{color}
\usepackage{hyperref}
\usepackage{amsmath}
\usepackage{amssymb}
\usepackage{algorithm}
\usepackage{algpseudocode}

\usepackage[T1]{fontenc}

\usepackage[utf8]{inputenc}

\usepackage{microtype}

\title{Content Reduction, Surprisal and Information Density Estimation for Long Documents}

\author{Shaoxiong Ji~\textsuperscript{1} \quad Wei Sun~\textsuperscript{2} \quad Pekka Marttinen~\textsuperscript{3} \\
\textsuperscript{1} University of Helsinki, Finland \quad
\textsuperscript{2} KU Leuven, Belgium \quad
\textsuperscript{3} Aalto University, Finland \\
  \texttt{shaoxiong.ji@helsinki.fi;sun.wei@kuleuven.be;pekka.marttinen@aalto.fi} \\
  }

\begin{document}
\maketitle
\begin{abstract}
Many computational linguistic methods have been proposed to study the information content of languages.
We consider two interesting research questions: 
1) how is information distributed over long documents, and 
2) how does content reduction, such as token selection and text summarization, affect the information density in long documents. 
We present four criteria for information density estimation for long documents, including surprisal, entropy, uniform information density, and lexical density. 
Among those criteria, the first three adopt the measures from information theory. 
We propose an attention-based word selection method for clinical notes and study machine summarization for multiple-domain documents. 
Our findings reveal the systematic difference in information density of long text in various domains.
Empirical results on automated medical coding from long clinical notes show the effectiveness of the attention-based word selection method.

\end{abstract}

\section{Introduction}
\label{sec:introduction}

\begin{figure}[htbp!]
\begin{center}
\begin{subfigure}[b]{0.5\textwidth}
	\includegraphics[width=\linewidth]{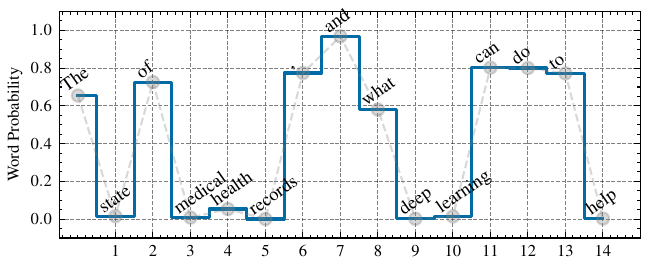}
	\caption{A text example and its word probability}
	\label{fig:example_sent}
\end{subfigure}
\qquad
\begin{subfigure}[b]{0.5\textwidth}
	\includegraphics[width=\linewidth]{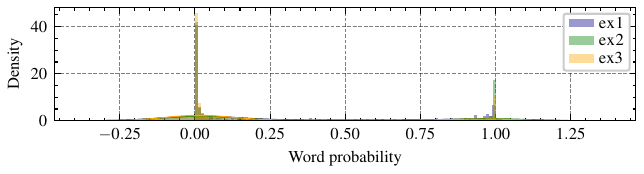}
	\caption{Examples of clinical notes and their word probability distributions}
	\label{fig:example_doc}
\end{subfigure}
\caption{Illustrations of a) a text snippet and its surprisal modeled by the pretrained language model, and b) word probability distributions of three examples of clinical notes}
\label{fig:example}
\end{center}
\end{figure}

Long document comprehension is an arduous task in human language understanding.
Information redundancy is becoming prevalent with the digitalization of individuals' records and the generation of massive user content.
Natural language encodes information with words and syntax. 
From the viewpoint of information theory~\citep{shannon1948mathematical}, language transmits information over a bandwidth-limited noisy channel.
Redundant information in long documents increases the cognitive load of readers, hinders the processing of texts, and probably affects the classification performance, especially for complex examples, in downstream domains. 
A rational language user tends to use information-dense phrases~\citep{levy2006speakers}.
The redundancy also increases the length of sequences, leading to extra computational costs for neural text encoders. 
Redundancy is linked to a reduced form of original content without sacrificing comprehension or cognition. 
For example, given a news categorization task, the sentence ``The state of medical health records, and what deep learning can do to help'', we can infer this category is about health and technology even with some key phrases such as ``medical health records'' and ``deep learning'' who have low word probabilities~(high surprisal) in Figure~\ref{fig:example_sent}. 

Information redundancy in long documents has been observed as a critical problem. 
Taking health text as an example, text redundancy in Electronic Health Records (EHR) has been widely recognized~\citep{wrenn2010quantifying}.
We illustrate the word probabilities of three examples of clinical notes in Figure~\ref{fig:example_doc} using a pretrained BERT base model~\citep{devlin2019bert}, where informative words and less informative words (tend to be redundant) are distributed at the two ends of the plots with large densities. 
Electronic clinical notes suffer from information redundancy mainly due to copy-and-paste in clinical notes.
Moreover, different expressions exist for the same thing in the clinical context. 
A study on 23,630 clinical notes shows that 46\% and 36\% of text are copied and imported, respectively~\citep{wang2017characterizing}.
Massive redundant information in clinical notes can lead to clinicians' burnout and increase medical coder working hours~\citep{montgomery2019burnout}. 
More worsening, it can lead to other harms, such as inconsistency in texts and error propagation during decision-making.

Human reading comprehension can be robust to errors and redundancy~\citep{hahn2019character}, while a robust neural text encoding model that can achieve human-level comprehension is a challenging research topic. 
Recent contextualized language models such as BERT (Bidirectional Encoder Representations from Transformers)~\citep{devlin2019bert} have achieved exciting performance on many natural language processing tasks. 
Like human comprehension from prior experience or education, pretrained models have prior exposure to specific training corpora. 
However, most of them are limited to processing the short sequence due to the quadratic complexity of the self-attention mechanism. 
For example, BERT is pretrained with a length of 128. 
Several efficient transformers have been proposed to solve the complexity issue to some extent~\cite {tay2022efficient}.
Nevertheless, an interesting question is how language models pretrained with short sequences transfer to long document representation learning. 

This paper studies the problem of long document encoding.
We view text understanding as an abstraction process. 
Specifically, we mimic the abstraction via two concrete text processing approaches, i.e., attention-based word selection and (abstractive) machine summarization.
We investigate two aspects: 
1) content reduction to shorten long documents via attention-based word selection and automated text summarization with pretrained models; 
2) information density estimation for original and content-reduced texts via a pretrained language model, surprisal model, entropy, uniform information density, and lexical density.
Our contributions are as follows.
\begin{itemize}
	\item We investigate the systematic difference in information density in different domains (i.e., clinical texts, movie reviews, and news articles) before and after content reduction. 
	\item We propose a simple pipeline-based method powered by the label attention mechanism to select informative words from lengthy clinical notes and perform automated medical coding.
	\item Our findings show that content reduction redistributes the information density of less standard text, such as clinical notes and movie reviews, and the information density reflects the downstream classification performance. Our empirical results also show that attention-based word selection can improve the performance of medical coding from clinical notes.
\end{itemize}

\section{Information Density Estimation}
\label{sec:information}

Similar to the definition of mass density in physics, information density in computational linguistics measures the human-readable information encoded per linguistic unit. 
One common metric to measure information density is the lexical density (Section~\ref{sec:lexical_density}), which describes the proportion of content words in a given corpus~\citep{kalinauskaite2018detecting}. 
Psycholinguistic experiments have shown a link between information density and other issues such as readability and memory~\cite{howcroft2017psycholinguistic}.
Generally, more grammatical words give less information, and lexical words such as nouns and verbs are more informative.  
We investigate the long document embeddings through the lens of information-theoretic estimation. 
Several measurements inspired by the information theory are adopted in this study, including surprisal model (Section~\ref{sec:surprisal}), entropy (Section~\ref{sec:entropy}), and uniform information density (Section~\ref{sec:UID}).

\subsection{Surprisal Model}
\label{sec:surprisal}

The surprisal model of human language processing describes the surprisal of a word given its prefix.
Intuitively, cognitive efforts involved with text understanding should be proportional to word surprisal.  
The lexical-based surprisal measure (Eq.~\ref{eq:surprisal}) in psycholinguistic evaluation~\citep{hale2001probabilistic,levy2008expectation} is defined as the negative logarithmic conditional probability given preceding words of $w_{k+1}$ (or its so-called context).
The surprisal is calculated as:
\begin{equation}
\label{eq:surprisal}
S=-\log P\left(w_{k+1} \mid w_{1} \ldots w_{k}\right),
\end{equation}
where $w_k$ is the $k$-th word.
The surprisal score values the amount of surprise. 
A higher surprisal value means a word difficult to process or comprehend. 
An error word should be more surprising than the correct word. 
For example, the misspelled word \textit{artial} and letter-transposed word \textit{atrila} produce more surprisal and are more difficult to comprehend than the correct word \textit{atrial}. 
\citet{demberg2013incremental} summarized two ways to estimate surprisal in psycholinguistic evaluation, i.e., lexical surprisal and structural surprisal. 
Lexical surprisal further considers two levels of word and part-of-speech, while structural surprisal depends on the syntax of sentence prefixes. 

Given a sentence $u=[w_1, \dots, w_t, \dots, w_n]$ and a pretrained contextualized language model such as BERT parameterized by $\boldsymbol{\theta}$, we can calculate the conditional probability of $t$-th word $w_t$ by applying the softmax transform on the $t$-th hidden representation $\mathbf{h}_t$ as
\begin{equation}
\label{eq:probability}
	p_{\boldsymbol{\theta}}\left(w_{t} \mid w_{<n}\right)=\operatorname{softmax}\left(\mathbf{W} \mathbf{h}_{t}+\mathbf{b}\right),
\end{equation}
where $\mathbf{W} \in \mathbb{R}^{|\mathcal{S}| \times d_{h}}$, $\mathbf{b} \in \mathbb{R}^{|\mathcal{S}|}$, and $|\mathcal{S}|$ is the vocabulary size of the target corpus $\mathcal{S}$. 
Accordingly, there are two approaches to computing sentence-level surprisal.
We use the n-gram model.
For example, the 3-gram model is defined as
\begin{align*}
	P(s)&=P\left(w_{1}\right) \times P\left(w_{2} \mid w_{1}\right) \times P\left(w_{3} \mid w_{2} w_{1}\right) \times  \\
	& \prod_{i=4}^{n} P\left(w_{n} \mid w_{n-1} w_{n-2} w_{n-3}\right).	
\end{align*}
Noise in text, like typos or errors, can degrade the context of a word, leading to increased surprise and increasing the difficulty of comprehension~\citep{hahn2019character}.
We investigate the surprisal level of texts from different domains, long documents, particularly to understand the behavior of neural text encoders. 

\subsection{Entropy}
\label{sec:entropy}

Entropy estimate has been studied in many ways.
\citet{genzel2002entropy} conducted a $n$-gram entropy estimate in three different ways, i.e., a $n$-gram probabilistic model, a probabilistic model induced by a statistical parser, and a non-parametric estimator.
The authors proposed the constancy rate principle governing language generation.
However, their local entropy estimate ignored the context. 
\citet{bentz2016word} used entropy to measure the average information content of natural languages and conducted a quantitive analysis to investigate the systematic difference in word entropies across different languages.
Inspired by these two works, we estimate entropy by utilizing pretrained contextualized language models to consider the context information and study the systematic difference in word entropies for long documents and their summaries across different domains.
The entropy of text $s$ is defined as
\begin{equation}
\label{eq:entropy}
H(s)=-\sum_{i=1}^{n} P\left(w_{i}\right) \log\left(P\left(w_{i}\right)\right),
\end{equation}
where $P(w_{i})$ is the probability of word $w_i$.
$P(w_{i})$ can be approximated as $P(w_{i})=\frac{f_{i}}{\sum_{j=1}^{n} f_{j}}$ from the frequency viewpoint and $f_i=\operatorname{freq}(w_i)$ is the frequency of word $w_i$.
Our study approximates it as the conditional probability generated by a pretrained language model given its context.

\subsection{Uniform Information Density}
\label{sec:UID}

The uniform information density (UID) hypothesis asserts that information encoding aims to transmit messages in a uniform way during the language production~\cite{jaeger2006redundancy,jaeger2010redundancy}.
The intuition behind UID is to maximize the information transmission and minimize comprehension difficulty. 
The UID hypothesis aligns with the principle of language production, i.e., to avoid information overloading or being uninformative.
The context plays an important role in the information density of sentences. 
If the context is considered, the information density of sentences is uniform; otherwise, it experiences an increase with the sentence number in local measures of entropy~\cite{genzel2002entropy}. 
\citet{meister2021revisiting} quantifies the linguistic uniformity by defining the UID as
\begin{equation}
	\mathrm{UID}^{-1}({u})=\frac{1}{n} \sum_{i=1}^{n} \Delta\left(S\left(u_{i}\right), \mu_{c}\right)
\end{equation}
where $\mu_c$ is an average information rate and $\Delta(\dot,~\dot)$ is a per-unit distance metric. 

From this viewpoint, UID can be regarded as a measure on how uniform the sentence conveys its meaning.
We investigate if the embeddings of long documents from pretrained language models adhere to the uniform information density hypothesis.

\subsection{Lexical Density}
\label{sec:lexical_density}

We first use lexical readability to examine how difficult a document is to understand. 
We apply the Flesch reading ease score that was introduced for reading ease evaluation~\citep{kincaid1975derivation}. 
It is formulated as:
\begin{equation*}
\scriptsize
206.835-1.015\left(\frac{\text { total words }}{\text { total sentences }}\right)-84.6\left(\frac{\text { total syllables }}{\text { total words }}\right),
\end{equation*}
where the coefficients come from user study. 
A higher score means easier to read.
A score of 100 indicates the text is effortless to read, while a score ranging from 0 to 10 means the text is complicated to comprehend and needs professional knowledge. 
We transfer the readability test to some specific domains and provide a reference for lexical density estimation. 

We study lexical richness, which basically measures to what extent different words are used in the text. 
Many lexical richness measures calculate the proportion of unique words to evaluate the lexical diversity~\citep{wimmer1999vocabulary}.
The widely used type-token ratio is calculated by the number of types divided by the number of tokens. 
We use one of its variants called the Herdan lexical richness measure proposed by \citet{herdan1960type}.
Herdan lexical richness is defined as: 
\begin{equation*}
	C=\frac{\log V(N)}{\log N},
\end{equation*}
where $N$ is the number of tokens and $V$ is the number of types. 

\section{Content Reduction}
\label{sec:content_reduction}

We cast long document understanding as a generation process that comprehends the long documents and digests key messages as latent states. 
As a result, the understanding process generates some short versions of the original text but preserves the subject matter of original long documents.
Specifically, we instantiate the generation process by two concrete instances, i.e., attention-based word selection and abstractive text summarization. 
Attention-based word selection in the previous section is similar to extractive text summarization. 
However, it is not trained with a reference dataset with extraction-based summaries.

\subsection{Attention-based Word Selection}
We propose a simple and efficient pipeline-based word selection method powered by the label attention mechanism that prioritizes essential information in the hidden representation relevant to medical codes. 
The label attention uses dot product to calculate the attention score matrix $\mathbf{A}\in \mathbb{R}^{n\times m}$ as:
\begin{equation}
\mathbf{A}=\operatorname{Softmax}(\mathbf{H} \mathbf{U}),
\end{equation}
where $\mathbf{H} \in \mathbb{R}^{n\times h}$ is the hidden features, $\mathbf{U}\in \mathbb{R}^{{h}\times m}$ is the parameter matrix of the query, and $m$ is the number of medical codes. 
We use mean pooling to obtain the attention vector $\mathbf{a}\in \mathbb{R}^n$ for word selection, i.e., $\mathbf{a}=\text{MeanPooling}(\mathbf{A})$.
Given a threshold or $q$-th quantile of the pooled attention score, we selected words whose attention scores meet the selection criteria, and other words in a text are filtered out.  
This pipeline can be extended to various text feature extractors that capture sequential dependency and utilize label-aware representations from the label attention mechanism.

\subsection{Text Summarization}
Automated text summarization transforms lengthy documents into shortened paragraphs while preserving the overall meaning. 
Abstractive summarization summarizes the text differently rather than extracting some key sentences from the document. 

We utilize two advanced abstractive summarization models, i.e., pretrained BART~\citep{lewis2020bart} that is trained by learning to predict the arbitrarily corrupted text and T5~\citep{raffel2020exploring} based on a text-to-text framework.
These two representative models have shown superior performance on several text summarization benchmarks.
However, there exists one limitation to this study. 
As the reference dataset with human summarization is not available, we can not tell which machine summarization model is the best. 

\section{Results and Analyses}
\label{sec:experiments}

\subsection{Tasks and Datasets}
\label{sec:datasets}
We conduct experiments on a long document classification task with three public datasets from different domains.
A statistical summary of datasets is shown in Table \ref{tab:data}.
\begin{table}[htbp!]
\centering
\small
\setlength{\tabcolsep}{2pt}
\begin{tabular}{l c c c c}
\toprule
Dataset & Avg. Length & Train & Validation & Test \\
\midrule
BBC News & 419 & 1,424 & 356& 445\\
IMDB & 698 & 1,553 & 350 & 1,791\\
MIMIC-III & 1,883 & 8,066 & 1,573 & 1,729\\	
\bottomrule
\end{tabular}
\caption{A statistical summary of datasets}
\label{tab:data}
\end{table}

\paragraph{Medical Coding}
Medical coding is a multi-label multi-class classification task that takes clinical notes from electronic health records as inputs and predicts medical codes of standard disease classification systems~\citep{ji2022unified}.
We use clinical notes from the MIMIC-III dataset~\citep{johnson2016mimic} and adopt the data split of top-50 codes from \citet{mullenbach2018explainable} that assigns frequently used ICD-9 codes to discharge summaries. 
We use the BERT text encoder as the neural backbone to get text representations and learn label-aware features with a label attention mechanism to boost the performance of medical coding.

\paragraph{News Topic Classification}
The BBC News dataset~\citep{greene2006practical} contains news articles in BBC News from 2004-2005.
It is used for news topic classification in five topical areas, i.e., business, entertainment, politics, sport, and technology.
Notice that we use our data split as there is no standard data partition. 

\paragraph{Movie Review Sentiment Analysis}
The IMDB movie review dataset~\citep{maas2011learning} has movie reviews posted on the IMDB website. 
As the average length of this dataset is relatively short, we select long reviews from the training and testing sets of the original data.
Then, we split an additional validation set for the IMDB long review data. 

\subsection{Results of Attention-based Selection for Medical Coding}
We present the results of the attention-based word selection method for medical coding. 
Specifically, we use this method to obtain selected texts whose sequence length is shorter than the original text. 
For the text encoder and code classifier, we adopt a recent medical coding model that utilizes recalibrated feature aggregation and multitask learning with focal loss~\citep{sun2021marn}.
After the word selection, we input the shortened text into a BERT-based medical coding model.
We compare this pipeline-based model with the following models.
The first category is the convolutional or recurrent neural network-based models. 
They are CAML~\citep{mullenbach2018explainable} that uses a text convolutional neural network and label attention mechanism, GatedCNN-NCI~\citep{ji2021medical} that adopts gated convolutions and a note-code interaction module, and JointLAAT~\citep{vu2020label} that utilizes bidirectional long short-term memory networks and a structured attention mechanism. 
The second category is based on BERT-based classifiers. 
We compare the truncated and hierarchical BERT~\citep{ji2021does} with three domain adaptive BERT models, i.e., PubMedBERT~\citep{gu2020domain}, BioBERT~\citep{lee2020biobert} and ClinicalBERT~\citep{alsentzer2019publicly}.
The third category is enhanced BERT-based models.
MDBERT~\citep{zhang2022hierarchical} considers three-level hierarchical encoding. 
Two variants of MDBERT are also compared. 
MDBERT-SBERT removes sentence BERT, and MDBERT+avg uses model ensemble. 
Our method achieves better performance than simple BERT-based classifiers and comparable performance than simple MDBERT without sentence BERT, although slightly worse than the ensemble-based method (MDBERT+avg).
Different word selection strategies also affect the performance of our method. 
Selection with the $q$-th quantile ($q=0.875$) is more flexible in selecting informative words and achieves better performance than the variant that only selects words with a fixed threshold for all documents.
We choose the threshold that can obtain a content-reduced text with an average length of 250.

\begin{table}[htbp!]
\small
\centering
\setlength{\tabcolsep}{3pt}
\begin{center}
\begin{tabular}{lrr|rr|r}
\toprule
 \multirow{2}{4em}{Model} 	& \multicolumn{2}{c}{AUC-ROC} & \multicolumn{2}{c}{ F1 } & \multirow{2}{2em}{P@5} \\  
 	&Macro &Micro& Macro&Micro &\\
\midrule
CAML	&	87.5	&	90.9	&	53.2	&	61.4	&	60.9	\\
GatedCNN-NCI	& 91.5 	&	93.8	&	62.9	 &	68.6	 &	65.3	 \\	
JointLAAT & 92.5 & 94.6 & 66.1& 71.6 & 67.1\\
\hline
PubMedBERT 	&	82.1	&	84.4	&	52.6	 &	57.3 	&	55.7	\\
BioBERT 	&	81.8	&	84.3	&	50.5	 &	55.4		&	54.5	\\ 
ClinicalBERT 	&	82.3	&	85.3	&	50.6	 &	56.9		&	55.7	\\
\hline
MDBERT-SBERT & 91.1 & 93.1 & 64.4 & 68.1 & 64.3 \\
MDBERT & 91.8 & 93.6 & 65.9 & 69.2 & 65.4 \\
MDBERT+avg & 92.8 & 94.6 & 67.2 & 71.7 & 67.4 \\
\hline
Ours (fixed) & 90.6	& 92.9	& 58.2	& 65.3	& 64.0 \\
Ours ($q$ quantile) & 91.6	& 93.5	& 64.6	& 68.5& 64.6 \\
\bottomrule
\end{tabular}
\end{center}
\caption{Attention word selection for medical coding on MIMIC-III dataset. ``+avg'' means averaging-based model ensemble. ``fixed'' and ``$q$ quantile'' denote that we select words by a fixed threshold and $q$ quantile.} 
\label{tab:selection_mimic3}
\end{table}%

\begin{figure}[ht!]
\centering
\includegraphics[width=0.3\textwidth]{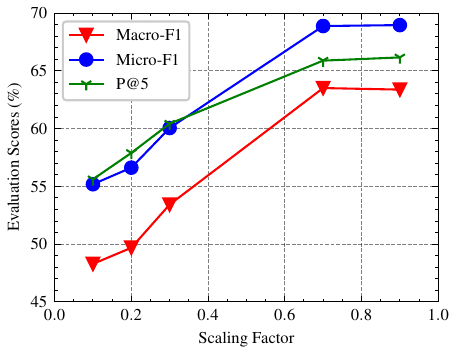}
\caption{Predictive performance on MIMIC-III dataset with  different scaling factors applied on the embeddings of selected words}
\label{fig:scaling_selected_words}
\end{figure}

We compare the model performance by applying a scaling factor to the embeddings of selected words to verify the effect of attention-based word selection further. 
Specifically, the word embeddings multiplied by the scaling factor is penalized as a restricted input signal to the model. 
Intuitively, we use the scaled word embeddings to control the strength of selected words.
A higher scaling factor means that the embeddings of selected words weigh more in the input text. 
Results in Figure~\ref{fig:scaling_selected_words} show that selected words with rich information are contributing more to representing the content, and the predictive performance is getting better with the increase of the scaling factor.
Then, we apply the scaling factor to both selected words and those not selected. 
Table~\ref{tab:selection_mimic3} shows the performance drops after applying a scaling factor of 0.1. 
The results reveal that penalizing the signal strength of selected words with the scaling factor leads to a more significant performance drop than downweighting those not selected words. 
These two studies indicate that attention-based word selection can extract important words to the model prediction. 

\begin{table}[htbp!]
\small
\centering
\setlength{\tabcolsep}{3pt}
\begin{center}
\begin{tabular}{lrr|rr|r}
\toprule
 \multirow{2}{4em}{Scaling} 	& \multicolumn{2}{c}{AUC-ROC} & \multicolumn{2}{c}{ F1 } & \multirow{2}{2em}{P@5} \\  
 	&Macro &Micro& Macro&Micro &\\
\midrule
$\text{Non-selected Words}$	& -2.9 	&	-2.1	&	-10.6	 &	-7.5	 &	-3.3	 \\	
$\text{Selected Words}$	&	-6.0	&	-5.0	&	-19.9	&	-16.6	&	-11.7	\\
\bottomrule
\end{tabular}
\end{center}
\caption{Performance drop on MIMIC-III dataset when applying a scaling factor of 0.1 on selected and non-selected words} 
\label{tab:selection_mimic3}
\end{table}%

\subsection{Results of Classification on Abstractive Summary}
Text summarization can be a way to alleviate the reader's workload by automatically extracting critical information from the text.
We test the performance of text classification on the abstractive summary. 
Experimental results show it is hard to achieve better performance on summaries than the original texts. 
In several cases, the performance drop can even be 10\%. 
We do not report those negative results in detail to avoid verbosity.   

\begin{figure*}[ht!]
\begin{center}
\begin{subfigure}[b]{0.3\textwidth}
	\includegraphics[width=\linewidth]{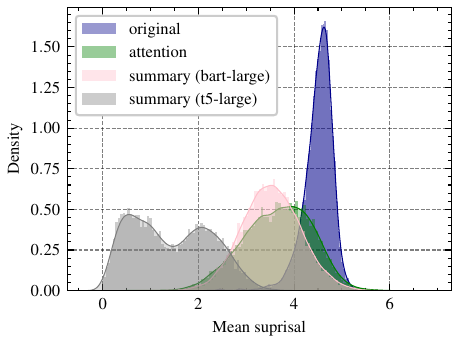}
	\caption{MIMIC-III}
	\label{fig:surprisal_mimic}
\end{subfigure}
\quad
\begin{subfigure}[b]{0.3\textwidth}
	\includegraphics[width=\linewidth]{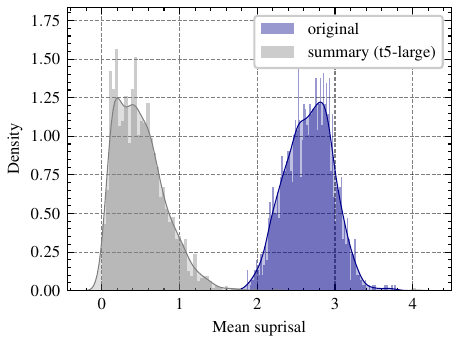}
	\caption{IMDB}
	\label{fig:surprisal_imdb}
\end{subfigure}
\quad
\begin{subfigure}[b]{0.3\textwidth}
	\includegraphics[width=\linewidth]{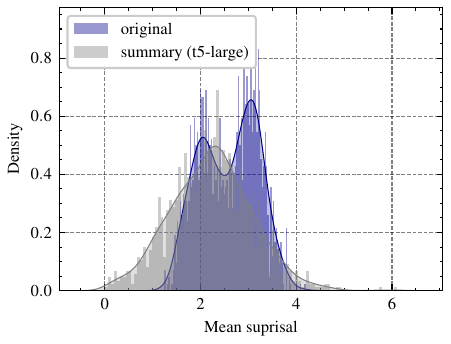}
	\caption{BBC News}
	\label{fig:surprisal_bbc_news}
\end{subfigure}
\caption{Kernel density estimation plots for document-level mean surprisal}
\label{fig:surprisal}
\end{center}
\end{figure*}

\subsection{Analysis of Surprisal, Entropy, and Uniform Information Density}

\paragraph{Surprisal}
We draw the kernel density estimation plot for document-level mean surprisal in Figure~\ref{fig:surprisal}.
The figure also shows histograms normalized to the same scale as the density curves.  
The smooth density estimation curve is generated by summing the Gaussians of individual points of word probabilities.  
The figures show that long original documents tend to have a higher mean surprisal than attention-selected texts and summaries. 
T5 summarization model significantly reduces the surprisal of MIMIC-III clinical notes and IMDB movie reviews. 

These results also align with the performance of downstream document classification tasks. 
With close surprisal distributions (Figure~\ref{fig:surprisal_bbc_news}), the downstream classification performance on the original text and summaries of BBC News is very close.
In contrast, the performance on IMDB summary drops up to 10\% when compared with the performance on original texts, which can partly be explained by the redistributed surprisal as shown in Figure~\ref{fig:surprisal_imdb}.

\begin{figure*}[ht!]
\begin{center}
\begin{subfigure}[b]{0.3\textwidth}
	\includegraphics[width=\linewidth]{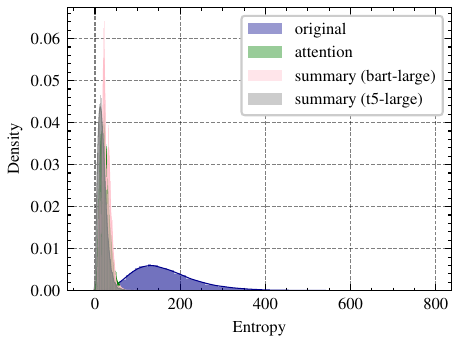}
	\caption{MIMIC-III}
	\label{fig:entropy_mimic}
\end{subfigure}
\quad
\begin{subfigure}[b]{0.3\textwidth}
	\includegraphics[width=\linewidth]{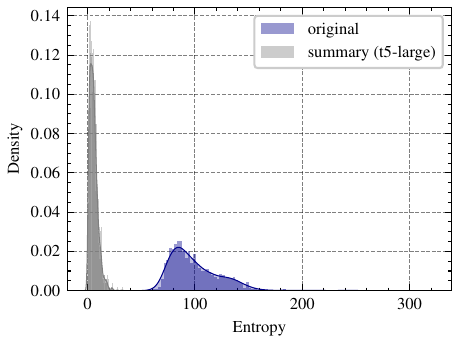}
	\caption{IMDB}
	\label{fig:entropy_imdb}
\end{subfigure}
\quad
\begin{subfigure}[b]{0.3\textwidth}
	\includegraphics[width=\linewidth]{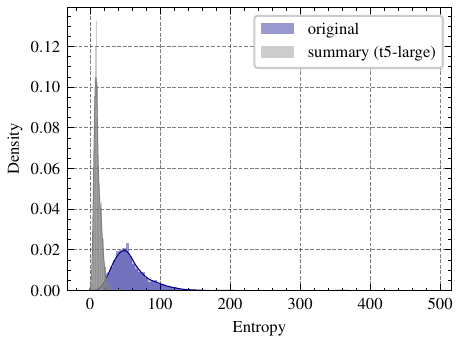}
	\caption{BBC News}
	\label{fig:entropy_bbc_news}
\end{subfigure}
\caption{Kernel density estimation plots for document-level entropy}
\label{fig:entropy}
\end{center}
\end{figure*}

\paragraph{Entropy}
Entropy describes the amount of information required to represent an event randomly drawn from the distribution. 
We estimate the document-level entropy to understand the information required to represent the text encoded by a language model.
Figure~\ref{fig:entropy} shows the kernel density estimation plots for document-level entropy of three datasets.
We can clearly see that the original texts contain more information than summaries of all three datasets and content-reduced text via attention selection in the MIMIC-III dataset. 

\paragraph{Uniform Information Density}

\begin{figure*}[htbp!]
\begin{center}
\begin{subfigure}[b]{0.23\textwidth}
	\includegraphics[height=2.8cm]{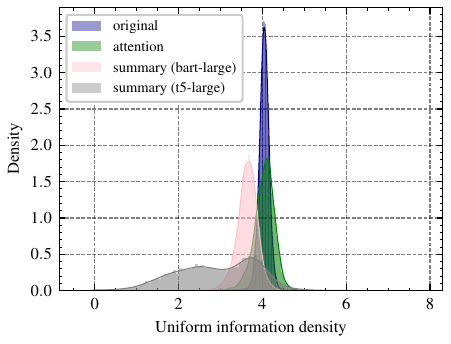}
	\caption{MIMIC-III}
	\label{fig:uid_mimic}
\end{subfigure}
\begin{subfigure}[b]{0.23\textwidth}
	\includegraphics[height=2.8cm]{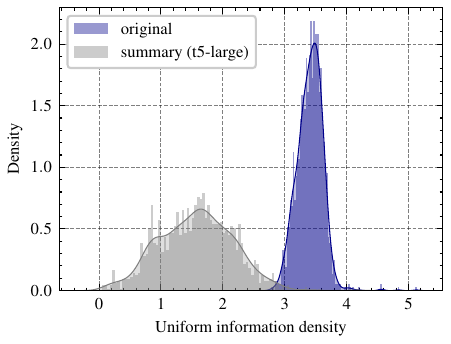}
	\caption{IMDB}
	\label{fig:uid_imdb}
\end{subfigure}
\begin{subfigure}[b]{0.23\textwidth}
	\includegraphics[height=2.8cm]{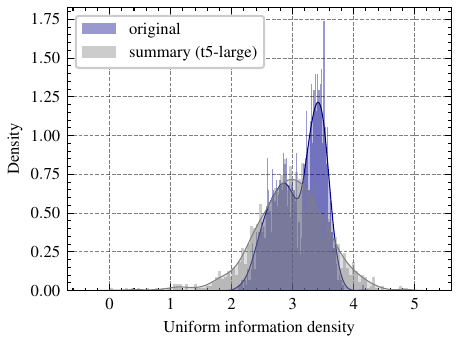}
	\caption{BBC News}
	\label{fig:uid_bbc_news}
\end{subfigure}
\begin{subfigure}[b]{0.23\textwidth}
	\includegraphics[height=2.8cm]{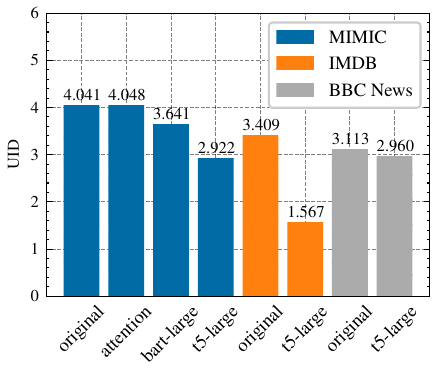}
	\caption{Corpus-level mean}
	\label{fig:uid_mean}
\end{subfigure}
\caption{Uniform information density. (a-c) document-level kernel density estimation plots and (d) mean UID in corpus level}
\label{fig:uid}
\end{center}
\end{figure*}

Uniform information density describes how uniformly the information is distributed through the communication channel. 
Figure~\ref{fig:uid} shows the kernel density estimation of UID of three datasets. 
In MIMIC-III, the UID distributions of the original text and attention-selected text overlap (Figure~\ref{fig:uid_mimic}), and the corpus mean UID of attention-selected text is slightly bigger than the original text.
Thus, we can conclude that attention selection can extract informative parts from the original text and maintain the level of uniformity. 
Figure~\ref{fig:uid_imdb} shows the summarization model generated IMDB review text with information density more uniformly distributed, while Figure~\ref{fig:uid_bbc_news} shows the overlapped UID of BBC News.
Figure~\ref{fig:uid_mean} of corpus-level UID further verifies summarization models generate texts with more uniform information density.

\subsection{Analysis of Lexical Density}
We conduct a quantitative analysis of the lexical structure of the original texts and content-reduced texts and summaries. 
Many metrics have been used to evaluate the complexity of language.
We choose two from those widely used metrics. 
They are lexical readability and lexical richness which measure how readable and diverse the text is.  
We conduct this analysis to examine the change in lexical complexity before and after the content reduction via attention selection and machine summarization. 

\begin{figure}[ht!]
\begin{center}
\begin{subfigure}[b]{0.22\textwidth}
	\includegraphics[width=\linewidth]{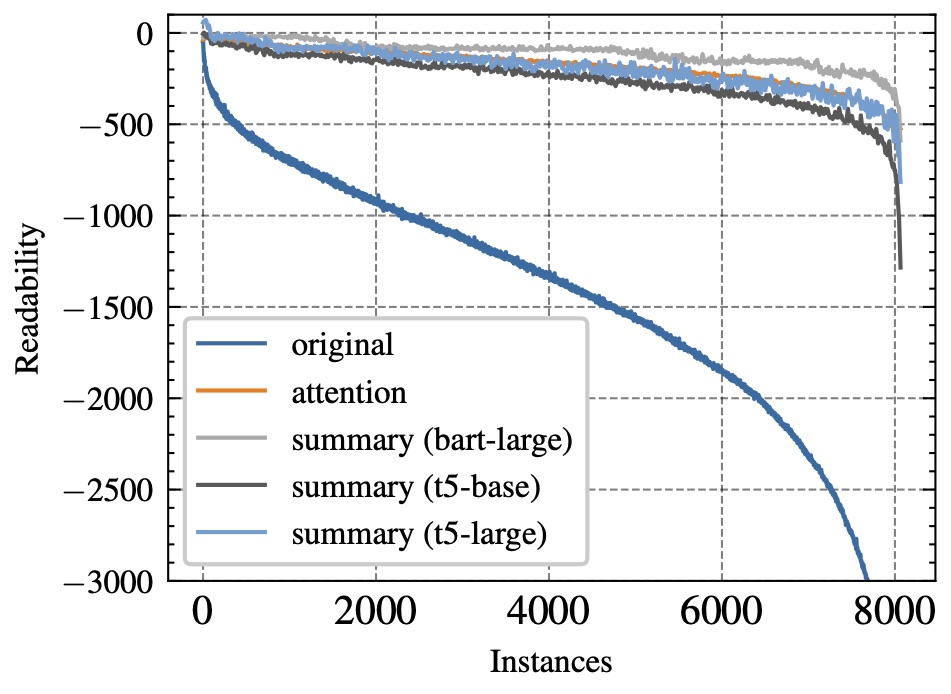}
	\caption{Readability of MIMIC}
	\label{fig:readability_mimic}
\end{subfigure}
\quad
\begin{subfigure}[b]{0.22\textwidth}
	\includegraphics[height=2.6cm]{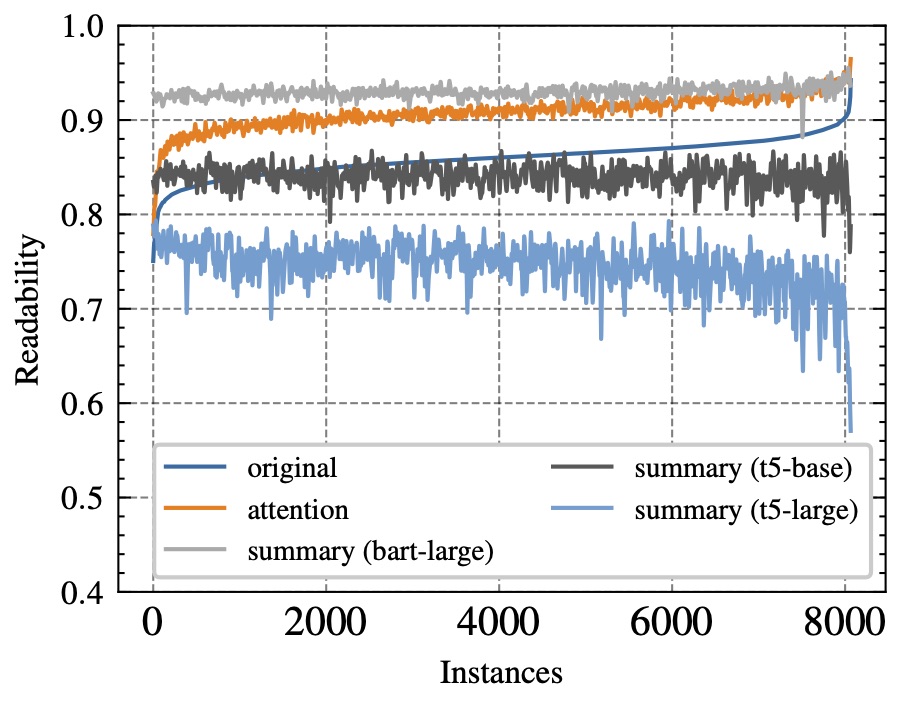}
	\caption{Richness of MIMIC}
	\label{fig:lexical_richness_mimic}
\end{subfigure}
\quad
\begin{subfigure}[b]{0.22\textwidth}
	\includegraphics[width=\linewidth]{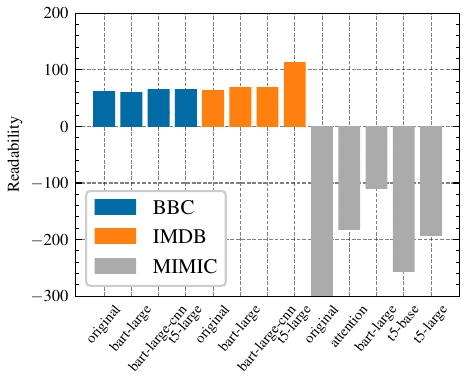}
	\caption{Mean readability}
	\label{fig:readability_corpus}
\end{subfigure}
\quad
\begin{subfigure}[b]{0.22\textwidth}
	\includegraphics[height=2.6cm]{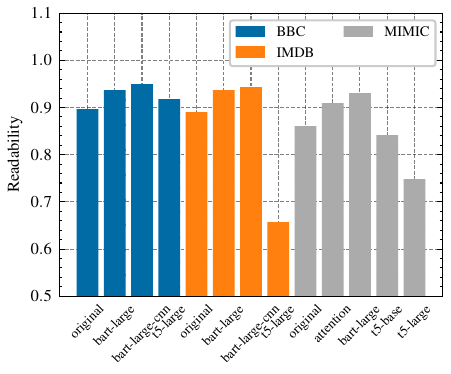}
	\caption{Mean richness}
	\label{fig:lexical_richness_corpus}
\end{subfigure}
\caption{Lexical readability measured by Flesch reading ease score and Herdan lexical richness of MIMIC-III at instance level and all three datasets at corpus level}
\label{fig:lexical_readability}
\end{center}
\end{figure}

We illustrate the lexical readability of each instance in the MIMIC-III dataset (Figure~\ref{fig:readability_mimic}) and the corpus mean of three datasets (Figure~\ref{fig:readability_corpus}) measured by the Flesch reading ease score.
We use the instance indices of sorted scores of the original dataset to plot each instance. 
Figure~\ref{fig:readability_mimic} shows that MIMIC-III clinical notes are extremely hard to comprehend with negative readability scores. 
With content reduction via attention-based selection and summarization, the readability scores of MIMIC-III clinical notes improve significantly but are still very hard to read. 
As for text in the domains of newswire and movie review, abstractive summarization generates texts that are slightly easier or the same as easy to read. 
Movie reviews contain several documents hard to read. 
Formal language in the news is more readable than clinical notes and user-generated movie reviews, which aligns with the finding of the Flesch test. 
The Flesch test tends to give a higher score for text with easy words.

Figure~\ref{fig:lexical_richness_mimic} and Figure~\ref{fig:lexical_richness_corpus} illustrate the Herdan lexical richness of each instance in the MIMIC-III dataset and all three datasets at the corpus level.
We can see that summarization with the BART model increases lexical richness by a considerable margin in IMDB and BBC News.
Furthermore, attention-based word selection improves the richness of MIMIC-III clinical notes (Figure~\ref{fig:lexical_richness_mimic}). 
T5 tends to generate less affluent summaries.

Content reduction increases the lexical density to some extent, especially for less standard text such as clinical notes. 
However, the behavior of different summarization models varies. 
Considering the performance boost brought by the attention-based word selection, we summarize that a simple pipeline method that condenses the lexical density of noisy texts (e.g., clinical notes) benefits the downstream classification task to some extent.
More investigation is needed for the text summarization method via transfer learning.

\subsection{Discussion, Limitations and Future Work}
This study sheds light on the investigation of an attention-based word selection method for clinical notes and information density estimation on various summarization texts. 
Extractive summarization by attention selection has shown good downstream performance on the medical coding task. 
However, some critical issues remain unexplored; for example, what if attention-based word selection filters out negations and breaks the syntactic structure?
Besides, statistical significance cannot be thoroughly evaluated given the limited number of training instances and domain data. 
We leave these unexplained problems for future work.
We tried some amendments for some existing limitations.
There are no ground-truth values of surprisal. 
An alternative we used is to approximate it via pretrained language models.
Also, we cannot directly evaluate the quality of word selection and summarization due to the lack of a reference dataset.
Instead, we evaluate it through downstream classification tasks.

\section{Related Work}
\label{sec:related}
Processing long documents with redundant information is burdensome. 
Many efforts have been made to estimate the redundancy in clinical notes and study the potential risks of redundancy in a retrospective manner. 
\citet{wrenn2010quantifying} quantified the redundancy in the clinical document by measuring the amount of new information and showed information duplication between document types. 
\citet{zhang2011evaluating} studied several methods for measuring the redundancy in clinical texts. 
In the clinical domain, vocabulary and errors are relatively rare compared with generic texts.
\citet{searle2021estimating} showed that clinical text is less efficient in encoding information than open-domain text from the perspective of information theory and observed that some clinical notes in the MIMIC database could be 97-98\% redundant.

\citet{levy2006speakers} investigated the possibility of uniformity maximization of information density through syntactic reduction. 
\citep{meister2021revisiting} revisited the uniform information density hypothesis and interpreted the hypothesis as the regression to the mean information of a language.  
Information density estimation is an important research task.
Several works have been done to automatically measure text information density, such as from the perspective of lexical and syntactic features~\citep{kalinauskaite2018detecting}. 
\citet{horn2013using} used open information extraction system to extract facts and applied factual density, calculated as the number of facts divided by the document size, to measure the informativeness of web documents.

\section{Conclusion}
\label{sec:conclusion}
Long document processing is challenging for many reasons, such as the difficulty of capturing long-term dependency and noises in the long text. 
This paper studies the encoding of long documents via information density estimation and empirical analyses on content reduction. 
We systematically show the difference in information density between original long documents and content-reduced texts. 
We improve the performance of automated medical coding by using selected words as inputs when compared with simple baselines that use the same neural backbone.
We validate that careful word selection can obtain words that can redistribute the distribution of word probability and entropy.
Our study takes a positive step towards understanding language model-based long document encoding.
\clearpage

\section*{Limitations}
As an empirical study, this paper did not outperform the state-of-the-art method, such as the ensemble-based hierarchical model~\citep{zhang2022hierarchical}. 
Our analyses focus on standard self-attention-based transformer networks and masked language models.
Recent efficient transformers and pretrained models with other language modeling objectives are not considered in this study.
We leave them in the future work.

\bibliography{references}
\bibliographystyle{acl_natbib}

\end{document}